\documentclass{article}

 \usepackage[dblblindworkshop, final]{neurips_2025}

\usepackage[utf8]{inputenc} 
\usepackage[T1]{fontenc}    
\usepackage{hyperref}       
\usepackage{url}            
\usepackage{booktabs}       
\usepackage{amsfonts}       
\usepackage{nicefrac}       
\usepackage{microtype}      
\usepackage{multirow}
\usepackage{graphicx}
\usepackage{xcolor}         
\usepackage{siunitx}    
\usepackage{subcaption} 
\usepackage{tcolorbox}
\tcbuselibrary{breakable,skins}

\title{Smaller Models, Smarter Rewards: A Two-Sided Approach to Process and Outcome Rewards}
\workshoptitle{Foundations of Reasoning in Language Models}

\author{
  Jan Niklas Groeneveld\thanks{The work of the paper was performed when he was an research internship under SAP Lab, in collaboration with Stanford Human-Centered AI Institution.} \\
University of California, Irvine\\
Irvine, CA 92697 \\
\texttt{jan.groeneveld@outlook.de} \\
  \And
  Xi Qin \\
  SAP Lab \\
  Palo Alto, CA 94304 \\
  \texttt{grace.qin@sap.com} \\
  \AND
  Alexander Schaefer \\
  SAP Lab \\
  Palo Alto, CA 94304 \\
  \texttt{alexander.schaefer01@sap.com} \\
  \And
  Yaad Oren \\
  SAP Lab \\
  Palo Alto, CA 94304 \\
  \texttt{yaad.oren@sap.com} \\
}

\begin{document}

\newcommand{\graceraw}[1]{{\color{orange}#1}}
\newcommand{\todo}[1]{{\color{blue}#1}}

\newcommand{\hidden}[1]{}  

\maketitle

\begin{abstract}
  Generating high-quality code remains a challenge for Large Language Models (LLMs).  
For the evolution of reasoning models on this task, reward models are a necessary intermediate step.  
These models judge outcomes or intermediate steps.  
Decoder-only transformer models can be turned into reward models by introducing a regression layer and supervised fine-tuning.  
While it is known that reflection capabilities generally increase with the size of a model, we want to investigate whether state-of-the-art small language models like the Phi-4 family can be turned into usable reward models blending the consideration of process rewards and outcome rewards.  

Targeting this goal, we construct a dataset of code samples with correctness labels derived from the APPS coding challenge benchmark. We then train a value-head model to estimate the success probability of intermediate outputs. Our evaluation shows that small LLMs are capable of serving as effective reward models or code evaluation critics, successfully identifying correct solutions among multiple candidates. Using this critic, we achieve over a 20\% improvement in the search capability of the most accurate code out of multiple generations.
\end{abstract}

\section{Introduction}

While several improvements have been achieved on reasoning-related tasks through techniques like chain-of-thought prompting, model improvement through bootstrapping (\cite{zelikman2022starbootstrappingreasoningreasoning}), tree-of-thought decoding (\cite{yao2023treethoughtsdeliberateproblem}), and explicit policy optimization through the tree (\cite{feng2024alphazeroliketreesearchguidelarge}), the first major breakthrough in reasoning capabilities came with OpenAI's o1 series and its successor o3, o4-mini and GPT-5 thinking.
The inner workings of these models are not publicly known, which made Deepseek-R1 (\cite{deepseekai2025deepseekr1incentivizingreasoningcapability}) the first public model to offer advanced reasoning capabilities.
This model was shortly followed by two other Chinese models, Kimi k1.5 (\cite{kimiteam2025kimik15scalingreinforcement}) and DAPO (\cite{yu2025dapoopensourcellmreinforcement}), which gave similar performance and offered more insights into the training and algorithms. 

When ChatGPT 3.5 was released in 2022, its success was enabled through a technique called ``Reinforcement Learning Through Human Feedback'' (RLHF).  
A dedicated reward model in the decoder-only structure learned human preferences, and the generative model was fine-tuned using this reward signal (\cite{ouyang2022traininglanguagemodelsfollow}). Instead of a classification head to predict the next token, their last layer often predicts just a single signal: the score they assign to the text up to that token (\cite{zhong2025comprehensivesurveyrewardmodels}).

When looking at the inner workings of reasoning models, all seem to share the same approach of generating a lengthy chain of thought before outputting the final result.
This chain of thought helps the models bridge the gap between the prompt and the potentially complex answer, providing helpful intermediate steps that extend the context used by the model to predict the tokens the answer is made of.
When there are mistakes or hallucinations in the reasoning trace, or intermediate generations are unstructured, this can easily derail a generation and lead to a wrong output.

Hence, the training process of a reasoning model aims to increase the probability of \textit{good} reasoning traces (those leading to a correct result), and decrease the probability of \textit{bad} traces.




Recent work on reward modeling distinguishes process reward models (PRMs) – which score intermediate steps – from outcome reward models (ORMs) – which score only final outputs (\cite{wang2023math,lightman2023let}). ORMs provide a single holistic score at the end of generation, while PRMs offer step-wise feedback, enabling fine-grained supervision and improving credit assignment and sample efficiency in multi-step tasks (\cite{choudhury2025process, cui2025process}). ORMs are simpler and cheaper but suffer from sparse signals and delayed credit assignment (\cite{cui2025process, wang2025towards}). However, PRMs require fine-grained labels during training—often needing costly human annotations for intermediate steps (\cite{cui2025process}). In contrast, ORMs typically demand many more episodes to learn, which can make RL impractical for complex tasks without auxiliary signals (\cite{wang2025towards,choudhury2025process}).

Recent work blends PRM and ORM signals or adopts hierarchical schemes. \cite{wang2025towards} introduce Hierarchical Reward Models (HRM), which combine coarse outcome and fine process scores, allowing later steps to override earlier mistakes. Other work like ThinkPRM (\cite{khalifa2025process}) and PathFinder-PRM (\cite{deep2025error}) focus on data-efficient PRM training and calibration at inference, enhancing the reward quality. In parallel, implicit reward models (LLM-as-judge to collect outcome feedback) like PRIME (\cite{cui2025process}) are explored to sidestep expensive PRM annotation.


Inspired by the strengths and drawbacks in PRMs and ORMs, we aim to build a versatile reward model that can serve both roles. This paper investigates how well fine-tuned decoder-only LLMs can act as reward models, specifically in the context of Python coding applications.
Our key contributions in this paper are the following:  
\begin{itemize}
    \item We replace the last layer of Phi-4 models by a single output with sigmoid function. 
    This architecture are successfully finetuned into a reward model in just two episodes, achieving promising results.
    
    \item We demonstrate that our trained reward model can serve as an effective lightweight critic and also a confident intermediate step evaluator, improving the correctness of Python code generation by reliably selecting correct rollouts by 20+\% (refer to Appendix \ref{performance_improve} for details). 
    \item We extensively analyze the model’s scoring of full rollouts and intermediate reasoning steps, uncovering valuable insights into its evaluation behavior. 
    
\end{itemize}

\section{Related Work}
An early work on tree-based decoding strategies was made by \cite{yao2023treethoughtsdeliberateproblem}.
The authors introduced the terminology ``Tree of Thought'', and evaluated a tree-based reasoning process on different reasoning tasks.
They reported an immense increase in success rates on the benchmarks they evaluated.
There are two main differences to our work.
First, they used a different level of granularity, where a ``thought'' is a for example sentence.
Our work, in comparison, uses tokens as the level of granularity.
Second, for judging intermediate steps, the authors prompted LLMs via a text interface.
This model also answered in natural language, and didn't output a single number.
Moreover, the team mainly evaluated on non-coding reasoning tasks, while our work is focused on coding challenges.

A follow-up paper on ``Tree of Thought'' written by \cite{feng2024alphazeroliketreesearchguidelarge} is about decoding with Monte Carlo Tree Search.
This paper uses a similar verdict mechanism to estimate the state value of an intermediate step, and uses this to do Monte Carlo Tree Search.
While the authors reported good results in their decoding, they did not discuss the performance of their value estimator.
We look into that topic and provide a comprehensive discussion.
Additionally, they used sentence-based granularity and focused on non-coding tasks.

Similar work was published by \citet{yu2024ovmoutcomesupervisedvaluemodels} who also proposed a decoder-based judge.
Their paper focused on introducing the concept called ``Outcome-supervised Value Model'', classifying it into the frameworks of outcome reward models and process reward models, the two most common ways of rewarding reasoning models.
They argued the value model approach to be superior over process-based rewarding, because the value estimation contains a forecast into the future, compared to process reward models rewarding correct steps in the past.
Like the previously discussed paper, they also used math datasets to assess their performance.

Other works like \cite{lightman2023letsverifystepstep} compare outcome reward models to process reward models and find superior performance for process reward models.
However, their approach for ORMs didn't judge intermediate steps, but sampled whole rollouts instead.

Anthropic's work (\cite{kadavath2022language}) sheds light on the self-evaluation capabilities and self-knowledge prediction of language models. By prompting with questions in a specific format, the models in study evaluate the probability “P(True)” that their answers are correct. Then crucially, the models are trained to perform prediction without reference to any specific proposed answer. The study found that larger models exhibit encouraging performance, calibration, and scaling for P(True) across a diverse range of benchmarks. It also reveals that models can effectively predict P(IK) and even partially generalize this prediction across different tasks.

Large Language Models (LLMs) exhibit powerful but opaque behaviors during next-word prediction. Traditional uncertainty estimation approaches typically focus on final outputs, overlooking the intermediate generation process.
This work by \cite{bigelow2024forking} introduces the Forking Tokens Hypothesis — the idea that certain individual tokens can cause significant shifts in the trajectory of text generation if selected during decoding.
To explore this, the authors propose Forking Paths Analysis, a method that tracks uncertainty dynamics across each generated token, rather than only the final ones.
Using GPT-3.5 and various tasks, they discover evidence of dramatic shifts in model behavior triggered by specific tokens, including seemingly minor ones like spaces or function words (e.g., “that” or “who”).
These findings highlight chaotic uncertainty patterns within LLMs and challenge static views of model confidence.

\section{Methods}
\label{sec:methods}
\subsection{Notation}
Let $S=(s_1, ..., s_l)$ be a sequence of tokens of length $l$, where each token comes from a token vocabulary $\Sigma$.
This sequence contains the prompt in Appendix \ref{appendix:prompts} sent to the generation LLM, and an arbitrary number of generated tokens.
The LLM uses this context to output a multinomial random distribution that assigns each token $s\in\Sigma$ a probability $p_{(s_1, ..., s_l)}(s) \in [0,1]$ to be the next token.\footnote{Of course, this probability distribution can have additional hyperparameters such as the temperature.}
In our situation, tokens are sampled from this distribution.
After the whole rollout was generated (either the end token was sampled, or the context window is fully used), the output can be judged and assigned a binary correctness label.
This judging happens by executing testcases given in the coding dataset on the generated code in a sandbox environment.
In formal notation, the judge calculates the following function:
\begin{equation}
    \mathcal{J}: \Sigma^* \rightarrow \left\{0, 1\right\}.
\end{equation}
The judge can only give its verdict once the whole code piece is submitted.
It can not decide whether intermediate steps are correct, a huge limitation when it comes to tree-based decoding.

\subsection{Dataset Generation}
\label{sec:dataset_generation}
Our goal is to estimate a state value $v_i$ for each position $i$, $1 \leq i \leq s_l$, i.e. the probability of sampling a correct result based on the prefix tokens $(s_1, ..., s_{i})$.
In formal math, let $\mathcal{G}(s_1, ..., s_{i})$ be the generation from the chosen prefix as a random variable.
With the binary judge $\mathcal{J}$, the state value is defined as
\begin{equation}
    v_i = \mathbb{E}\left[J\left(\mathcal{G}(s_1, ..., s_{i})\right)\right].
\end{equation}

APPS contains both problems that expect input and output via I/O operations, and function completion with a designated function interface. They mostly come with test cases to evaluate the generated code. We use a slightly different prompting framework for these two types of problems, with generation prompts listed in the appendix \ref{appendix:prompts}. However, calculating an accurate ground truth vector with all exhaustive solutions for each problem is computationally infeasible.
Instead, we generate 36 rollouts per problem setup. Later on, we would use this dataset of rollouts labeled by ground-truth correctness for model training and evaluation.
To hold these rollouts representative for a tree-based case, these 36 rollouts are not generated independently from the prompt.
Instead, we start with one main rollout (generated from Microsoft's Phi-4-mini), select six branching positions, and then generate six rollouts from each branching position, resulting in $6 \times 6 = 36$.

Let $p_i$ be a short notation for $p_{s_1, ..., s_{i-1}}(s_i)$, the probability the model assigned to the $i$-th token with respect to the context $(s_1, ..., s_{i-1})$ in the main rollout.
Tokens that are part of the prompt receive probability $1$ to have a unified notation for prompt-tokens and generated tokens.
When looking at the observed distribution of these $p_i$s in many tasks, the vast amount of probabilities is very close to one, and a few tokens have low probability.
For these probabilities to be low, there are two possible explanations.
First, these positions could be really ambiguous, with many synonyms as options, or important decisions for the following reasoning.
Second, there could be a token with high probability, but just a different path is taken, because sampling is used, and the decision fell for an unlikely token.
In both cases, these low-probability tokens are the important branch positions for our reasoning path.
Branching here will likely result in different outputs, while branching at high-probability token will likely share a common sequence with the previous rollout. Utilizing this pattern, we select a set $\mathcal{I}$ with $n_b=\left|\mathcal{I}\right|=6$ indices as the set of the positions with the lowest probabilities.
For each of these positions $i\in\mathcal{I}$, we generate $k=6$ rollouts, and evaluate the generated code on the unit tests given in the APPS dataset by \citet{apps:hendrycksapps2021}.

Therefore, our training dataset is a collection of coding problem description, sample solution, system prompt, and reasoning traces, we provide one relatively short example in Appendix \ref{appendix:sample_data}. 

\subsection{Training}
In our experiments, we train and test value estimations based on the 3.8B Phi-4-mini and 14B Phi-4 model over two episodes.  
At a high level, we perform fine-tuning on the regression head plus several last layers of these models. Table \ref{table:general_information_training_data} and Appendix \ref{appendix:hyperparameter} provide details of standard hyperparameters and number of last layers.  

To estimate the state values as needed, we take the same Phi-4-mini and Phi-4 14B architectures and replace the classifier layer at the network output with a linear regression layer. This modification yields a decoder-only architecture that predicts the value of a partial reasoning trace at a given point, based solely on preceding tokens. The regression layer is designed to estimate the probability of success from each token onward, while maintaining the causal constraint of using only past context. Since the output is probabilistic, we apply a sigmoid transformation to the regression output and train the model using binary cross-entropy loss. For value estimation, we used a batch size of 64 (for the 14B model) and 24 (for the 3.8B model), and fine-tuned the last 12 layers for both models at a learning rate of $1\mathrm{e}{-4}$. 

After fine-tuning, our first analysis objective is whether the trained models are capable as an ORM, which are able to distinguish successful rollouts from unsuccessful ones.  
Also, we take a look into the models' capability to judge intermediate reasoning steps. 

\begin{table}[h]
\centering
\begin{tabular}{p{6cm}|p{1.5cm}}
\toprule
Number of problems in test dataset & 465 \\
Number of problems in train dataset & 3,984 \\
total number of problems & 4,449 \\
\midrule
Train dataset size unbalanced & 66,924 \\
Train dataset size balanced & 110,016 \\
\midrule
Batch size Phi-4-mini-instruct (4B) & 64 \\
Finetuned Layers Phi-4-mini-instruct (4B) & 236-248 \\
\hline
Batch size Phi-4 (14B) & 24 \\
Finetuned Layers Phi-4 (14B) & 180-192 \\
\hline
Optimizer & Adam \\
Learning Rate & 1e-4 \\
\bottomrule
\end{tabular}
\caption{
    Key data on datasets and training hyperparameters
}
\label{table:general_information_training_data}
\end{table}

\subsection{Imbalanced and Balanced Dataset}
When Phi-4-mini-instruct generates the rollouts for the problems, the ratio of correct and incorrect rollouts is usually imbalanced.
While many problems have more incorrect than correct rollouts, other problems have a higher number of correct rollouts, and the fraction of correct rollouts is typically closely related to the difficulty of the problem for the model.
Whenever we train on the raw, imbalanced dataset, we open the door to misleading interpretations and a few biases the model could pick up.
First, with an imbalanced dataset the accuracy becomes a less useful metric, especially if the class imbalance is not reported.
Therefore, testing on the imbalanced version of the dataset comes with caveats.
Second, seeing the distribution of correct and incorrect rollouts per problem can enable the model to just learn the \textit{difficulty of the problem statement}, and only little about the correctness of the reasoning.
Suppose there are four correct and 32 incorrect rollouts for a problem.
The sequences have the same prefix (the problem statement), and differ later.
If the model just learns to predict ``false'' for this problem statement or this type of problem, it will minimize the loss fairly well.
To do so, it is enough to pick up shallow features like complexity, length, or formatting of the problem statement while ignoring the reasoning trace or encoding.
So, the reward model can learn to estimate the difficulty of a problem as a proxy for its correctness probability.
That enables it to ``hack'' the accuracy metric without ever learning to judge the actual results.

Therefore, we also train and test our reward models on a balanced dataset.
To create this balanced dataset, for each problem we oversample the smaller group (correct/incorrect) to match the number of correct and incorrect rollouts per problem.
This is done for both the train and test dataset.
With this addition, we take away the option for the model to just learn the difficulty of the problem, and the accuracy numbers are more interpretable.

\section{Results \& Discussions}
\label{sec:exp_results}
\paragraph{Classification Performance}
\begin{table*}[h]
\centering
\sisetup{table-align-text-post=false} 
\begin{tabular}{
  p{1em} 
  l      
  S[table-format=2.1] 
  S[table-format=2.1] 
  S[table-format=2.1] 
  S[table-format=2.1] 
}
\toprule
 & \textbf{Model} & \multicolumn{2}{c}{\textbf{3.8B Phi-4-mini-instruct}} & \multicolumn{2}{c}{\textbf{14B Phi-4}} \\
\cmidrule(lr){3-4} \cmidrule(lr){5-6}
& \textbf{Balanced / Imbalanced Training} & {\textbf{Balanced}} & {\textbf{Imbalanced}} & {\textbf{Balanced}} & {\textbf{Imbalanced}} \\
\midrule
\multirow{8}{*}{\rotatebox{90}{\textbf{Imbalanced Test Data}}} &
Predicted $>$ 0.5 & 33.5\,\% & 53.8\,\% & 46.3\,\% & 51.9\,\% \\
& Predicted $<$ 0.5 & 66.5\,\% & 46.2\,\% & 53.7\,\% & 48.1\,\% \\
\cmidrule(lr){2-6}
& Accuracy & 64.0\,\% & 66.0\,\% & 73.8\,\% & 71.7\,\% \\
\cmidrule(lr){2-6}
& If predicted $>$ 0.5, rollout correct & 65.2\,\% & 60.0\,\% & 69.8\,\% & 65.7\,\% \\
& If predicted $<$ 0.5, rollout incorrect & 63.3\,\% & 73.2\,\% & 77.2\,\% & 78.2\,\% \\
& If rollout correct, prediction $>$ 0.5 & 47.2\,\% & 72.2\,\% & 72.5\,\% & 76.5\,\% \\
& If rollout incorrect, prediction $<$ 0.5 & 78.3\,\% & 61.0\,\% & 74.8\,\% & 67.9\,\% \\
\midrule
\multirow{8}{*}{\rotatebox{90}{\textbf{Balanced Test Data}}} &
Predicted $>$ 0.5 & 31.4\,\% & 53.4\,\% & 45.6\,\% & 51.8\,\% \\
& Predicted $<$ 0.5 & 68.4\,\% & 46.6\,\% & 54.4\,\% & 48.2\,\% \\
\cmidrule(lr){2-6}
& Accuracy & 55.3\,\% & 51.9\,\% & 65.8\,\% & 60.5\,\% \\
\cmidrule(lr){2-6}
& If predicted $>$ 0.5, rollout correct & 58.5\,\% & 51.8\,\% & 67.3\,\% & 60.1\,\% \\
& If predicted $<$ 0.5, rollout incorrect & 53.9\,\% & 52.0\,\% & 64.5\,\% & 60.9\,\% \\
& If rollout correct, prediction $>$ 0.5 & 36.8\,\% & 55.2\,\% & 61.4\,\% & 62.3\,\% \\
& If rollout incorrect, prediction $<$ 0.5 & 73.9\,\% & 48.5\,\% & 70.2\,\% & 58.7\,\% \\
\midrule
\cmidrule(lr){2-6}
& Pass@1 on best $3 \choose 1 $, baseline 45\% & 50\% & 52\% & 55\% & 54\% \\
& Pass@3 on best $3 \choose 10$, baseline 65\% & 73\% & 72\% & 77\% & 78\% \\
\bottomrule
\end{tabular}
\caption{
    Comparison of the predictive performance of the Phi-4-mini based value prediction and the (regular) Phi-4 performance on a wide variety of metric, measured both on an imbalanced and balanced version of the test data.
    The baseline passing rates of pass@1, pass@3, and pass@10 are 45\%, 65\%, and 84\% are obtained without interference of any of the reward models, just the generation of Phi-4-mini-instruct.
}
\label{table:model_comparison}
\end{table*}

For training and evaluation, we use the APPS dataset, a combination of several public coding datasets.
APPS contains both problems that expect input and output via IO operations, and function completion with a designated function interface.
They mostly come with test cases to evaluate generated code.\footnote{For some problems, only few test cases are given, so the dataset has the unfortunate limitation that false-positives are possible.} As explained in section \ref{sec:dataset_generation}, we generate one base rollout for each of these samples, select forking tokens (six in our case), and generate also six rollouts from there.
Table \ref{table:model_comparison} shows the classification performance at a glance.
After training on the same datasets, the 14B model performs substantially better than the 3.8B Phi-4-mini, with the majority of prediction results achieving 60+\%. We observe 20+\% as the highest improvement base on the calculations in Appendix \ref{performance_improve}, where Pass@1 on best $3 \choose 1$ has the baseline of classification accuracy as 45\% and the highest as 22.2\%; Pass@3 on best $10 \choose 3$ has the baseline classification accuracy as 65\% and the highest as 20.0\%.

\begin{figure}[htbp]
    \centering
    
    \begin{subfigure}{0.48\textwidth}
        \includegraphics[width=\linewidth]{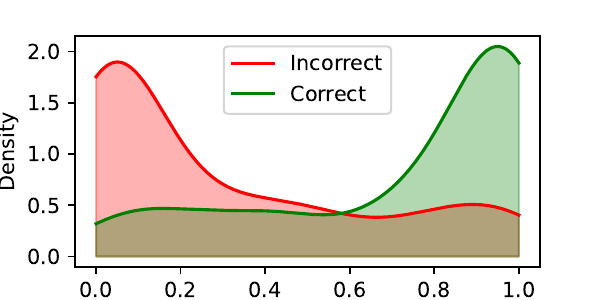}
        \caption{The trained 14B Phi-4 model's differentiation capability of the correct and incorrect rollouts on the test dataset. The correctness is based on the ground-truth verifier built with unit tests in a safe execution sandbox. }
        \label{fig:comparison-mini-normal}
    \end{subfigure}
    \hfill
    \begin{subfigure}{0.48\textwidth}
        \includegraphics[width=\linewidth]{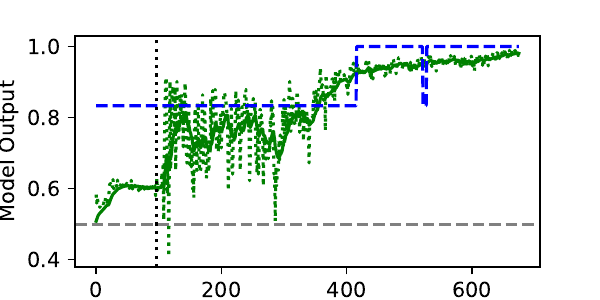}
        \caption{The trained 14B Phi-4 model's predicted belief over number of available tokens in a correct rollout as the green dotted line. The estimated ground truth is the blue dashed line, which is retrieved by the correct fraction in six forking positions. 
        }
        \label{fig:one_sample_trajectory}
    \end{subfigure}
    \caption{
        The CDF plot on the left is generated through kernel density estimation with Gaussian kernels. Then this model is used to generate the success probability estimations on the right.
    }
\end{figure}

The predictive performance of the 14B model is illustrated in Figure \ref{fig:comparison-mini-normal}.
The distribution shows the approximated probability to visualize how often a certain success probability estimation occurs for that group.

\paragraph{Performance on Balances vs Imbalanced Dataset}
As Table \ref{table:model_comparison} shows, the accuracies of the models on the imbalanced dataset are similar with minor differences.
However, on the balanced test dataset, we do see differences between the two types of model training, with the model trained on the balanced dataset performing substantially better.
However, when it comes to determining a relative order of rollouts, when the model needs to select the best or the best three rollouts, we see close accuracies between the two types of training.
In conclusion, these results suggest that balancing a train dataset has impact on the final performance, but it is rather limited.

In what follows, we discuss the key research questions (RQ) we pose and the answers our work offers.
\paragraph{RQ1: Can we use this reward model as a handy critic for code generation?}
Now with this trained reward model, we want to validate its capability to select correct rollouts out of multiple options for a problem from Table \ref{table:model_comparison}.
There are two key terminologies involved: 1. Pass@k is the number of passes of $k$ unit tests with the specific rollout; 2. Best ${m \choose n}$ is to select the best $n$ rollouts using the reward model out of $m$ rollouts.
\begin{figure}[htbp]
    \centering
    \includegraphics[width=0.8\linewidth]{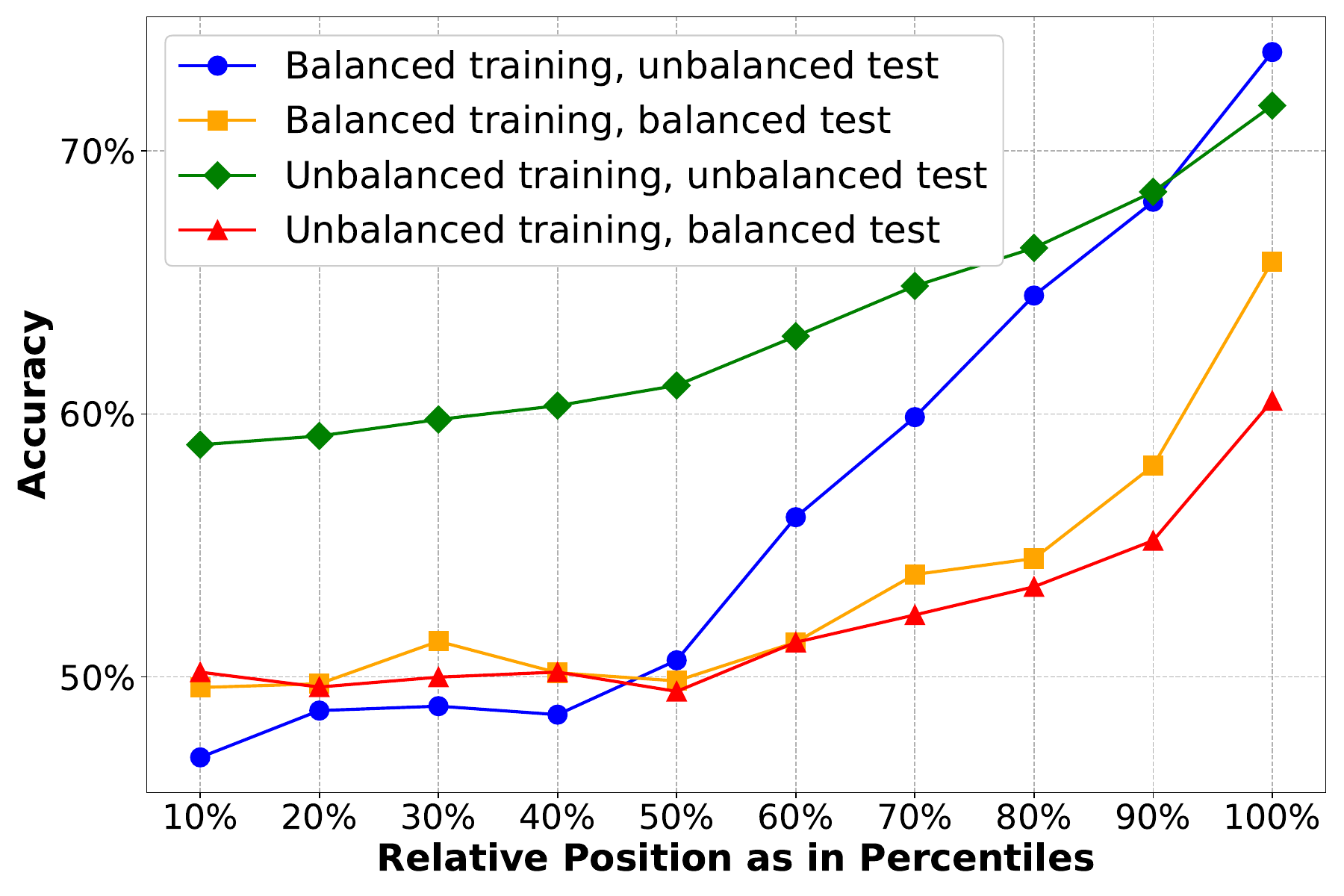}
    \caption{Accuracy of percentiles in rollout generation from four combinations of the 14B model trained and tested on balanced/imbalanced data.}
    \label{fig:by_position_accuracy_all}
\end{figure}
We observed that before applying the critic, pass@1 is 45\%, pass@3 is 65\%, and pass@10 is 84\%.
After applying the critic and measuring best ${m \choose n}$, we observe significant pass@1 improvement to 50\%–55\%, and pass@3 to 72\%–78\%.
Therefore, our findings support a confident affirmative answer to the research question: employing our model in a lightweight role as an evaluation critic or data quality filter enhances the correctness of Python code generation.

\paragraph{RQ2: Can this outcome reward model also be capable of judging intermediate generation steps?}
All the metrics above refer to the model's ability to provide a verdict on entire rollouts, i.e., the outcome reward.
To validate how often the reward model succeeds in estimating the probability of intermediate steps, we show percentile-accuracy plots such as Figure \ref{fig:by_position_accuracy_all}.
We measure the model's accuracy vs. the amount of given partial code, i.e., the accuracy of up to 100\% tokens from the first code line equals the accuracy in Table \ref{table:model_comparison}.
We argue that the earlier the model achieves higher accuracies, the more useful it is for early stopping or rollout guidance.
We observe that the model requires at least 50\% partial code to judge better than a random guess.

The decoder-only reward model estimates success probability after each token, enabling confidence tracking during rollouts.
Figure \ref{fig:one_sample_trajectory} shows a correct generation where the model’s confidence rises after seeing the output, aligning well with ground truth.
We observe a warming-up stage before around 100 tokens.
This is because our prompt asks the generation model to think in a Chain-of-Thought manner: first elaborate on the problem and describe a solution, and then write the code.
Given that the model waits to improve its performance until roughly the 50\% percentile, we suspect that the model might be able to catch errors in the code, but lacks the self-correction capability for already incorrect reasoning steps.
With Figures \ref{fig:by_position_accuracy_all} and \ref{fig:one_sample_trajectory}, we also confidently answer RQ2 in the affirmative.

The model begins to show meaningful performance (better than random guesses) as PRMs after approximately 50\% of the token generation process, with accuracy improving beyond the initial baseline. This indicates a lower bound of the preceding code tokens required to provide the predictive power.


\section{Conclusion}

These experiments demonstrate that, large-scale capacity is not a prerequisite for reward modeling: models from the Phi-4 family (14 B parameters) can serve as effective reward models, even when compared with state-of-the-art systems like GPT-5 and Claude 4 Sonnet at vastly larger scales of hundreds to trillions of parameters.
These models are capable of truly identifying errors in the rollouts.  
Moreover, they demo the capability of selecting correct rollouts, substantially increasing the probability of selecting a correct rollout.  
However, we see that model size is a primary limitation for this task, as the 3.8B Phi-4-mini-instruct falls significantly behind the 14B model.  

\section{Limitations}
Due to computational resource constraints, there are several experiments that should become our next steps as follows. 
\begin{itemize}
    \item Computing an accurate ground-truth vector for each problem is infeasible for the main solution variants and branching factors. Our current setup of generating 36 rollouts per problem already requires over 48 hours for one pass on a 4×A100 Azure computation instance. This throughput is mainly throttled by the output token throughput of the generation model Phi-4 mini. As a result, tuning the hyperparameters for branching decisions is left to future work.  
    \item During rollout generation, we exclude samples where the Phi model fails to produce any correct rollout. Our current method depends on a ground-truth correct solution such as the tree trunk, from which branching occurs. This cold start problem remains a challenge.
    \item Preliminary experiments with smaller branching factors (e.g., 6x6) indicate a positive impact from increased branching. However, we were unable to identify the point of diminishing returns, as broader branching was not computationally tractable in our current setup.
    \item APPS dataset comes with its own train-test split. Specifically, both their training and test datasets contain 5,000 samples. However, as our appendix \ref{apps_distribution} shows, the default test dataset follows an utterly different distribution. Therefore, we performed train-test-split on the original training set.
    \item Our current setup has the strong assumption that the distribution of the predicted correctness probability does not shift between the base policy and the policy in post-training. And we leave the exploration of post-training dynamics to future work.
\end{itemize}

\section{Acknowledgment}
We gratefully acknowledge Kanishk Gandhi of Stanford University Human-Centered Artificial Intelligence Institution, for his guidance in the literature review process and constructive contributions to the development of our methodology. We also attribute the implementation of the code evaluation sandbox to Gradyn Nagle, who was on rotation in our team at SAP during the development of this work.
\bibliography{reasoning_tree_paper}

@misc{zelikman2022starbootstrappingreasoningreasoning,
      title={STaR: Bootstrapping Reasoning With Reasoning}, 
      author={Eric Zelikman and Yuhuai Wu and Jesse Mu and Noah D. Goodman},
      year={2022},
      eprint={2203.14465},
      archivePrefix={arXiv},
      primaryClass={cs.LG},
      url={https://arxiv.org/abs/2203.14465}, 
}

@misc{yao2023treethoughtsdeliberateproblem,
      title={Tree of Thoughts: Deliberate Problem Solving with Large Language Models}, 
      author={Shunyu Yao and Dian Yu and Jeffrey Zhao and Izhak Shafran and Thomas L. Griffiths and Yuan Cao and Karthik Narasimhan},
      year={2023},
      eprint={2305.10601},
      archivePrefix={arXiv},
      primaryClass={cs.CL},
      url={https://arxiv.org/abs/2305.10601}, 
}

@misc{feng2024alphazeroliketreesearchguidelarge,
      title={Alphazero-like Tree-Search can Guide Large Language Model Decoding and Training}, 
      author={Xidong Feng and Ziyu Wan and Muning Wen and Stephen Marcus McAleer and Ying Wen and Weinan Zhang and Jun Wang},
      year={2024},
      eprint={2309.17179},
      archivePrefix={arXiv},
      primaryClass={cs.LG},
      url={https://arxiv.org/abs/2309.17179}, 
}

@misc{yu2024ovmoutcomesupervisedvaluemodels,
      title={OVM, Outcome-supervised Value Models for Planning in Mathematical Reasoning}, 
      author={Fei Yu and Anningzhe Gao and Benyou Wang},
      year={2024},
      eprint={2311.09724},
      archivePrefix={arXiv},
      primaryClass={cs.AI},
      url={https://arxiv.org/abs/2311.09724}, 
}

@misc{lightman2023letsverifystepstep,
      title={Let's Verify Step by Step}, 
      author={Hunter Lightman and Vineet Kosaraju and Yura Burda and Harri Edwards and Bowen Baker and Teddy Lee and Jan Leike and John Schulman and Ilya Sutskever and Karl Cobbe},
      year={2023},
      eprint={2305.20050},
      archivePrefix={arXiv},
      primaryClass={cs.LG},
      url={https://arxiv.org/abs/2305.20050}, 
}

@misc{deepseekai2025deepseekr1incentivizingreasoningcapability,
      title={DeepSeek-R1: Incentivizing Reasoning Capability in LLMs via Reinforcement Learning}, 
        author={DeepSeek-AI and Daya et al. },
      
      year={2025},
      eprint={2501.12948},
      archivePrefix={arXiv},
      primaryClass={cs.CL},
      url={https://arxiv.org/abs/2501.12948}, 
}

@misc{kimiteam2025kimik15scalingreinforcement,
      title={Kimi k1.5: Scaling Reinforcement Learning with LLMs}, 
      author={Kimi Team and Angang et al.},
      year={2025},
      eprint={2501.12599},
      archivePrefix={arXiv},
      primaryClass={cs.AI},
      url={https://arxiv.org/abs/2501.12599}, 
}

@misc{yu2025dapoopensourcellmreinforcement,
      title={DAPO: An Open-Source LLM Reinforcement Learning System at Scale}, 
      author={Qiying Yu et al.},
      year={2025},
      eprint={2503.14476},
      archivePrefix={arXiv},
      primaryClass={cs.LG},
      url={https://arxiv.org/abs/2503.14476}, 
}

@article{bigelow2024forking,
  title={Forking paths in neural text generation},
  author={Bigelow, Eric and Holtzman, Ari and Tanaka, Hidenori and Ullman, Tomer},
  journal={arXiv preprint arXiv:2412.07961},
  year={2024}
}

@article{kadavath2022language,
  title={Language models (mostly) know what they know},
  author={Kadavath, Saurav and Conerly, Tom and Askell, Amanda and Henighan, Tom and Drain, Dawn and Perez, Ethan and Schiefer, Nicholas and Hatfield-Dodds, Zac and DasSarma, Nova and Tran-Johnson, Eli and others},
  journal={arXiv preprint arXiv:2207.05221},
  year={2022}
}

@article{apps:hendrycksapps2021,
  title={Measuring Coding Challenge Competence With APPS},
  author={Dan Hendrycks and Steven Basart and Saurav Kadavath and Mantas Mazeika and Akul Arora and Ethan Guo and Collin Burns and Samir Puranik and Horace He and Dawn Song and Jacob Steinhardt},
  journal={NeurIPS},
  year={2021}
}

@misc{ouyang2022traininglanguagemodelsfollow,
      title={Training language models to follow instructions with human feedback}, 
      author={Long Ouyang and Jeff Wu and Xu Jiang and Diogo Almeida and Carroll L. Wainwright and Pamela Mishkin and Chong Zhang and Sandhini Agarwal and Katarina Slama and Alex Ray and John Schulman and Jacob Hilton and Fraser Kelton and Luke Miller and Maddie Simens and Amanda Askell and Peter Welinder and Paul Christiano and Jan Leike and Ryan Lowe},
      year={2022},
      eprint={2203.02155},
      archivePrefix={arXiv},
      primaryClass={cs.CL},
      url={https://arxiv.org/abs/2203.02155}, 
}

@misc{zhong2025comprehensivesurveyrewardmodels,
      title={A Comprehensive Survey of Reward Models: Taxonomy, Applications, Challenges, and Future}, 
      author={Jialun Zhong and Wei Shen and Yanzeng Li and Songyang Gao and Hua Lu and Yicheng Chen and Yang Zhang and Wei Zhou and Jinjie Gu and Lei Zou},
      year={2025},
      eprint={2504.12328},
      archivePrefix={arXiv},
      primaryClass={cs.CL},
      url={https://arxiv.org/abs/2504.12328}, 
}

@article{choudhury2025process,
  title={Process reward models for llm agents: Practical framework and directions},
  author={Choudhury, Sanjiban},
  journal={arXiv preprint arXiv:2502.10325},
  year={2025}
}

@article{cui2025process,
  title={Process reinforcement through implicit rewards},
  author={Cui, Ganqu and Yuan, Lifan and Wang, Zefan and Wang, Hanbin and Li, Wendi and He, Bingxiang and Fan, Yuchen and Yu, Tianyu and Xu, Qixin and Chen, Weize and others},
  journal={arXiv preprint arXiv:2502.01456},
  year={2025}
}

@article{wang2025towards,
  title={Towards Hierarchical Multi-Step Reward Models for Enhanced Reasoning in Large Language Models},
  author={Wang, Teng and Jiang, Zhangyi and He, Zhenqi and Tong, Shenyang and Yang, Wenhan and Zheng, Yanan and Li, Zeyu and He, Zifan and Gong, Hailei},
  journal={arXiv preprint arXiv:2503.13551},
  year={2025}
}

@inproceedings{lightman2023let,
  title={Let's verify step by step},
  author={Lightman, Hunter and Kosaraju, Vineet and Burda, Yuri and Edwards, Harrison and Baker, Bowen and Lee, Teddy and Leike, Jan and Schulman, John and Sutskever, Ilya and Cobbe, Karl},
  booktitle={The Twelfth International Conference on Learning Representations},
  year={2023}
}

@article{wang2023math,
  title={Math-shepherd: Verify and reinforce llms step-by-step without human annotations},
  author={Wang, Peiyi and Li, Lei and Shao, Zhihong and Xu, RX and Dai, Damai and Li, Yifei and Chen, Deli and Wu, Yu and Sui, Zhifang},
  journal={arXiv preprint arXiv:2312.08935},
  year={2023}
}

@article{khalifa2025process,
  title={Process reward models that think},
  author={Khalifa, Muhammad and Agarwal, Rishabh and Logeswaran, Lajanugen and Kim, Jaekyeom and Peng, Hao and Lee, Moontae and Lee, Honglak and Wang, Lu},
  journal={arXiv preprint arXiv:2504.16828},
  year={2025}
}

@article{deep2025error,
  title={Error Typing for Smarter Rewards: Improving Process Reward Models with Error-Aware Hierarchical Supervision},
  author={Deep Pala, Tej and Sharma, Panshul and Zadeh, Amir and Li, Chuan and Poria, Soujanya},
  journal={arXiv e-prints},
  pages={arXiv--2505},
  year={2025}
}
\bibliographystyle{plainnat}


\newpage
\section*{Appendix}



\subsection{Prompts}
\label{appendix:prompts}

\begin{figure}[htbp]
    \centering
    \begin{tcolorbox}[
        colback=black!5!white, 
        colframe=black!75!white, 
        breakable, 
        arc=4mm, 
        boxsep=2mm, 
        left=2mm, right=2mm, top=2mm, bottom=2mm 
    ]
    <|system|>You are a Python programmer in a programming competition. In the following, you are given a coding challenge.
    
    Please write code to solve the coding challenge. Follow the formatting instructions for input and output exactly.
    
    Get inputs via input(), and write outputs via print().
    
    Before writing the code, do the following:
    
    1. Elaborate on the problem. Highlight the key challenges
    
    2. Write the core algorithm idea in at most three sentences
    
    3. Elaborate on the idea in more detail.
    
    The code must be a directly executable script, not a function.
    
    NEVER include example usage or hard-coded data!
    
    ALWAYS generate the full code!

    Here is an example for a very simple problem:
    
    \verb|```|python
    
    number\_of\_testcases = int(input())
    
    for \_ in range(number\_of\_testcases):
    
    \,\,\,\,a, b, c = input().split()
        
    \,\,\,\,b = b[::-1]
        
    \,\,\,\,if a in b and b in c:
        
    \,\,\,\,\,\,\,\,print("TRIPLE TWIST")
            
    \,\,\,\,else:
        
    \,\,\,\,\,\,\,\,print("NO")
            
    \verb|```|
    
    <|end|><|user|>\{task\}<|end|><|assistant|>
    \end{tcolorbox}
    \caption{The prompt we use for problems that require I/O operations. We use the chat template which the model was trained with, laying out both system and user prompts. In the system prompt, we give instructions and give an piece of example code. The example code is for a straightforward string problem that is neither in the test nor train dataset. The model is supposed to continue with its thoughts and its code after the assistant tag.}
    \label{fig:prompt_io}
\end{figure}

\begin{figure}[htbp]
    \centering
    \begin{tcolorbox}[
        colback=black!5!white, 
        colframe=black!75!white, 
        arc=4mm, 
        boxsep=2mm, 
        left=2mm, right=2mm, top=2mm, bottom=2mm 
    ]
    <|system|>You are a Python programmer in a programming competition. In the following, you are asked to write code for a function.
    
    At the beginning of the function, do the following in the docstring:
    
    1. Elaborate on the problem. Highlight the key challenges
    
    2. Write the core algorithm idea in at most three sentences
    
    3. Elaborate on the idea in more detail.

    NEVER include example usage!
    
    ALWAYS generate the full code!
    
    <|end|><|user|>\{task\}<|end|><|assistant|>
    \verb|```|python
    
    def \{function\_signature\}
    
    \,\,\,\,
    """
    \end{tcolorbox}
\caption{The prompt we use for problems that require a complete function as the solution. In the assistant part of the message, we directly give the function signature to force the model to produce a valid function, and also not deviate from the given signature. Moreover, we force the model to start with a docstring. Using this prompt, we significantly reduced the number of malformed generations.}
    \label{fig:prompt_function}
\end{figure}

\newpage

\subsection{Sample Data}
\label{appendix:sample_data}
Please refer to Figure \ref{fig:sample_data} as one example of training data sample.
\begin{figure}[htbp]
    \centering
    \begin{tcolorbox}[
        colback=black!5!white, 
        colframe=black!75!white, 
        breakable, 
        arc=4mm, 
        boxsep=2mm, 
        left=2mm, right=2mm, top=1mm, bottom=1mm 
    ]
    <|system|>You are a Python programmer in a programming competition. 
    
    In the following, you are given a coding challenge. Please write code to solve the coding challenge. Follow the formatting instructions for input and output exactly. Get inputs via input(), and write outputs via print(). Before writing the code, do the following: 1. Elaborate on the problem. Highlight the key challenges 2. Write the core algorithm idea in at most three sentences 3. Elaborate on the idea in more detail. The code must be a directly executable script, not a function. NEVER include example usage or hard-coded data! ALWAYS generate the full code! Here is an example for a very simple problem: 
    
    \verb|```|python 
    
    number\_of\_testcases = int(input()) for \_ in range(number\_of\_testcases): 
    
    \,\,\,\,a, b, c = input().split() 
    
    \,\,\,\,b = b[::-1] 
    
    \,\,\,\,if a in b and b in c: 
    
    \,\,\,\,\,\,\,\,print("TRIPLE TWIST") 
    
    \,\,\,\,else: 
    
    \,\,\,\,\,\,\,\,print("NO")
    \verb|```|
    
    <|end|><|user|>You are given a permutation $p_1, p_2, \dots, p_n$. Recall that sequence of $n$ integers is called a permutation if it contains all integers from $1$ to $n$ exactly once. Find three indices $i$, $j$ and $k$ such that: $1 \le i < j < k \le n$; $p_i < p_j$ and $p_j > p_k$. Or say that there are no such indices. 
    
    -----Input----- 
    
    The first line contains a single integer $T$ ($1 \le T \le 200$) — the number of test cases. Next $2T$ lines contain test cases — two lines per test case. The first line of each test case contains the single integer $n$ ($3 \le n \le 1000$) — the length of the permutation $p$. The second line contains $n$ integers $p_1, p_2, \dots, p_n$ ($1 \le p_i \le n$; $p_i \neq p_j$ if $i \neq j$) — the permutation $p$. 
    
    -----Output----- 
    
    For each test case: if there are such indices $i$, $j$ and $k$, print YES (case insensitive) and the indices themselves; if there are no such indices, print NO (case insensitive). If there are multiple valid triples of indices, print any of them. 
    
    -----Example----- 
    
    Input 3 4 2 1 4 3 6 4 6 1 2 5 3 5 5 3 1 2 4 
    
    Output YES 2 3 4 YES 3 5 6 NO
    
    <|end|><|assistant|>
    
    The problem requires us to find a triplet of indices in a given permutation such that the values at those indices first increase and then decrease. The key challenges are to efficiently find such a triplet in each permutation, which can be done by iterating through the permutation and checking for the increasing and then decreasing condition. The core algorithm idea is to iterate through the permutation and maintain a candidate for the second element of the triplet (the peak). When we find a new maximum, we check if the previous element is less than the current one (forming the increasing part) and if the next element is less than the current one (forming the decreasing part). If both conditions are met, we have found our triplet. The idea in more detail is to iterate through the permutation once, keeping track of the current maximum and its index. If we find a new maximum, we check the elements before and after it to see if they form the increasing and decreasing parts of the triplet. If they do, we print the indices and break out of the loop. If we finish the loop without finding such a triplet, we print "NO". Here is the code: 
    
    \verb|```|python 

    T = int(input()) for \_ in range(T): 
    
    \,\,\,\,n = int(input()) 
    
    \,\,\,\,p = list(map(int, input().split())) 
    
    \,\,\,\,found = False 
    
    \,\,\,\,for i in range(1, n - 1): 
    
    \,\,\,\,\,\,\,\,if p[i - 1] < p[i] > p[i + 1]: 
    
    \,\,\,\,\,\,\,\,\,\,\,\,print("YES") 
    
    \,\,\,\,\,\,\,\,\,\,\,\,print(i, i + 1, i + 2) 
    
    \,\,\,\,\,\,\,\,\,\,\,\,found = True 
    
    \,\,\,\,\,\,\,\,\,\,\,\,break 
    
    \,\,\,\,\,\,\,\,if not found: 
    
    \,\,\,\,\,\,\,\,\,\,\,\,print("NO")
    
    \verb|```|
    
    \end{tcolorbox}
    \caption{Example of one piece of training sample.}
    \label{fig:sample_data}
\end{figure}


\subsection{Hyperparameters \& Hardware}
\label{appendix:hyperparameter}
In the stabilized experiment, we used a batch size of 64 for the 14B model and 24 for
the 3.8B model, at a learning rate of $1 \times 10^{-4}$. During the ablation, we probed batch sizes around 8, 12, 24, 64 to test the memory boundary in our Azure instance. We also tried learning rate at $1 \times 10^{-5}$ and $1 \times 10^{-6}$. 

Our computation instance type is Azure's Standard\_NC96ads\_A100\_v4 with 96 CPU cores (AMD EPYC 7V13), 880 GB RAM, 256 GB disk, and 4 Nvidia A100 GPUs in 320GB GPU memory. The total GPU hours are around 95.

\subsection{Performance Improvements}
\label{performance_improve}
From Table \ref{table:model_comparison}:
\begin{itemize}
    \item Pass@1 on best $3 \choose 1$
    \begin{itemize}
        \item Baseline classification accuracy is 45\%. 
        \item The ceiling of improvement is $ \frac{55-45}{45} \simeq 22.2\% $
    \end{itemize}
    \item Pass@3 on best $10 \choose 3$ 
    \begin{itemize}
        \item Baseline classification accuracy is 65\%.
        \item The ceiling of improvement is $ \frac{78-65}{65} \simeq 20\% $
    \end{itemize}
\end{itemize}

\subsection{APPS Data Distribution Study}
\label{apps_distribution}



\begin{figure}[htbp]
\centering
    \begin{subfigure}{0.48\textwidth}
        \includegraphics[width=\linewidth]{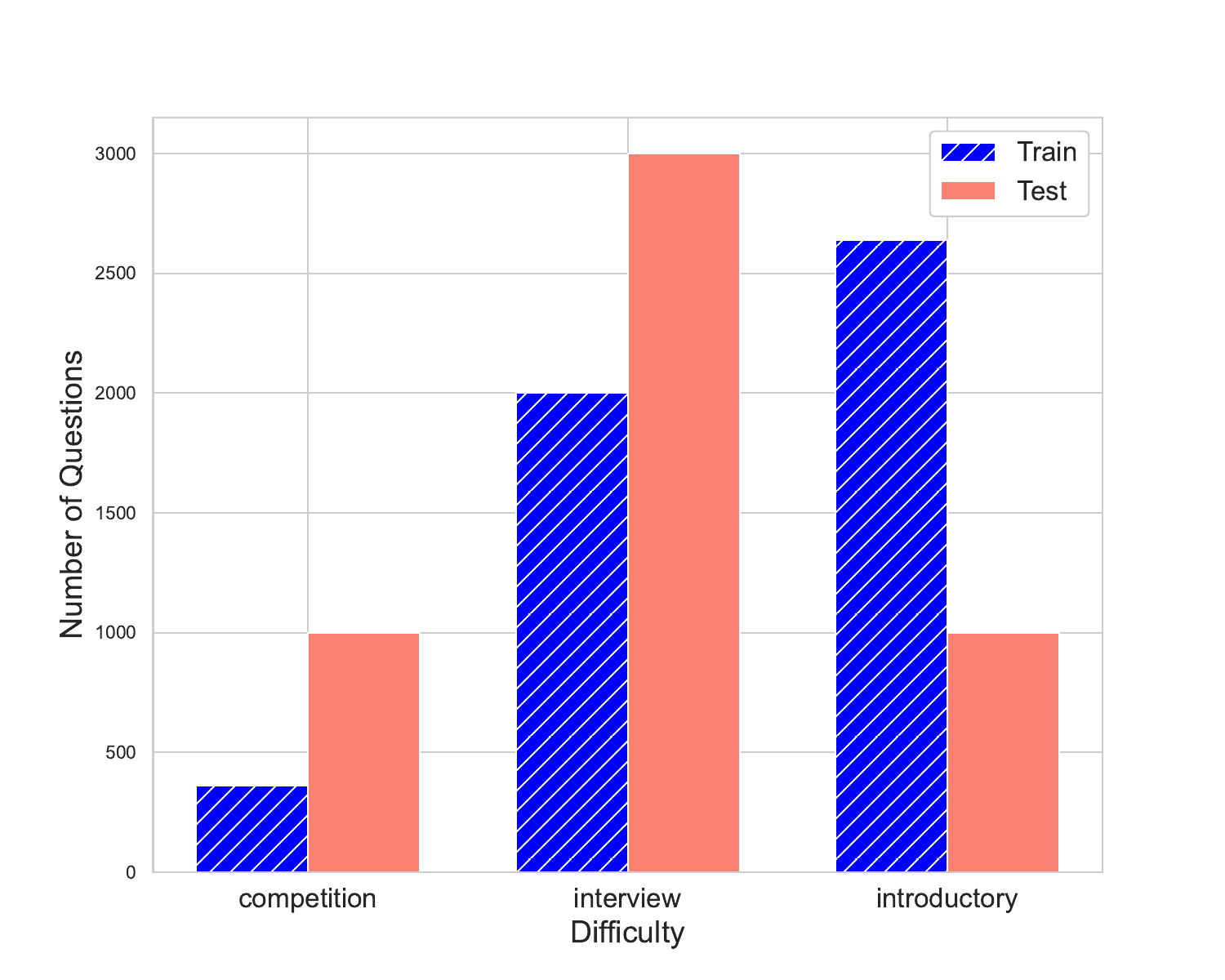}
        \caption{ The difficulty levels, which are from APPS dataset original definition. }
        \label{fig:bar}
    \end{subfigure}
    \hfill
    \begin{subfigure}{0.48\textwidth}
        \includegraphics[width=\linewidth]{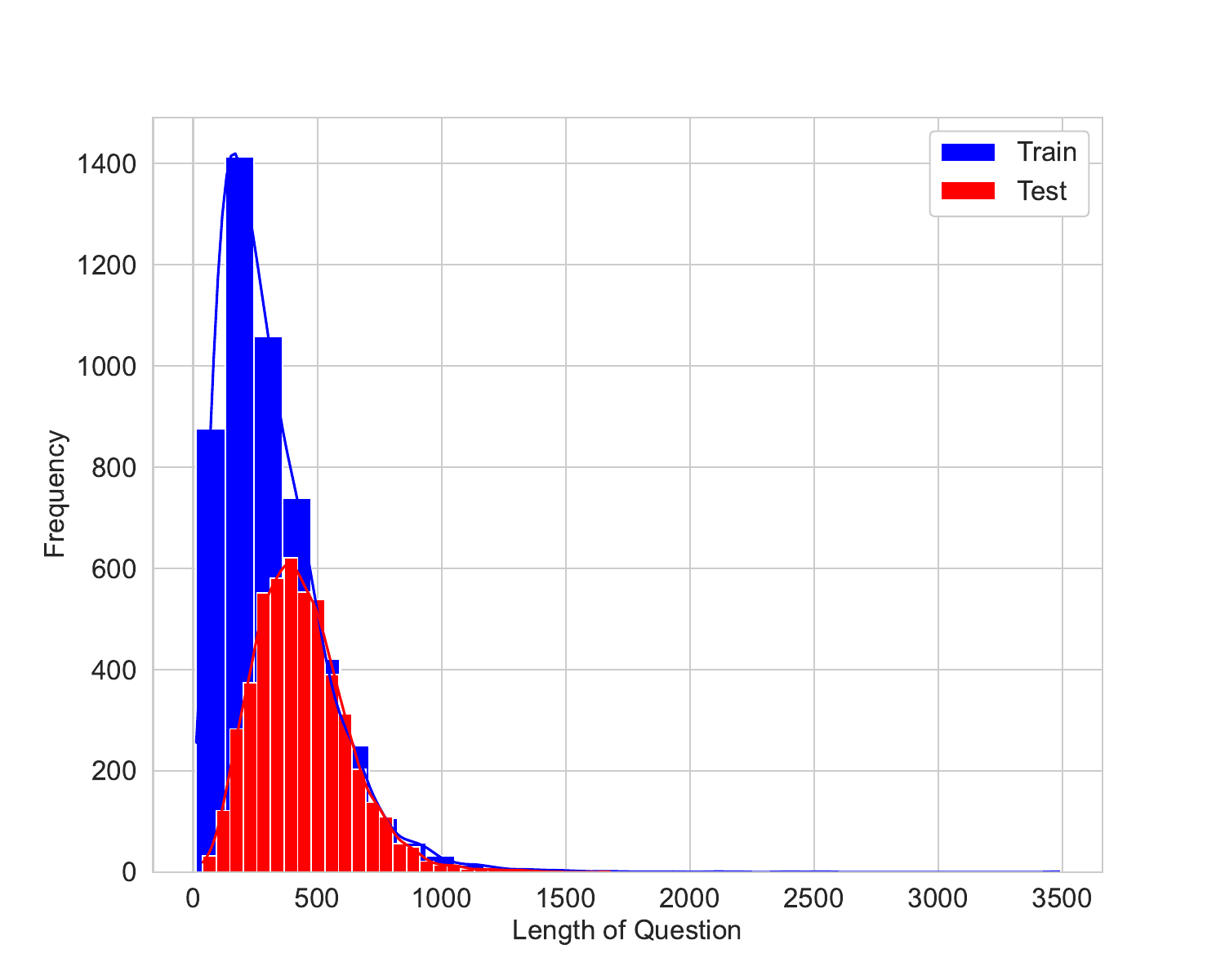}
       \caption{The length of question, which counts the token length per problem's question.}
        \label{fig:hist}
    \end{subfigure}


\label{fig:apps-data-distribution}
\end{figure}

We use the APPS Dataset, which is distributed under the MIT License. Our use complies with the license terms, and we will include the original copyright notice in our future artifacts to be released. In the paper where the APPS dataset was released, \cite{apps:hendrycksapps2021} highlighted differences in data quality, particularly in terms of unit test coverage, between the "train" and "test" splits. In addition to this observation, we want to emphasize that the intrinsic distribution of problem formats and complexity levels also varies significantly between these splits, as illustrated in Figure~\ref{fig:bar} and Figure~\ref{fig:hist}. These differences suggest that the train and test sets may not be fully aligned in terms of problem representation. This will substantially affect our results, as our method is designed to operate on the distribution of contextual tokens rather than the surface-level structure or complexity of the problem descriptions. Therefore, our method is more sensitive to such mis-alignment in data composition. We'll take the deep-dive of the performance impact from task formulation in APPS dataset to future work.

\end{document}